\documentclass[10pt,twocolumn,letterpaper]{article}

\usepackage{cvpr}              
\definecolor{cvprblue}{rgb}{0.21,0.49,0.74}
\usepackage[pagebackref,breaklinks,colorlinks,allcolors=cvprblue]{hyperref}
\usepackage{hyperref}
\usepackage{url}
\usepackage{array}
\usepackage{tabularx}

\usepackage{wrapfig}
\usepackage{graphicx}
\usepackage{booktabs}
\usepackage{multirow}

\def\confName{CVPR}
\def\confYear{2027}

\title{Image Diffusion Models Are Good 3D Human Motion Generators}

\author{
Siyuan Yang\textsuperscript{1*}\quad
Chong Wang\textsuperscript{2*}\quad
Hao Cheng\textsuperscript{3}\quad
Jun Liu\textsuperscript{4}\\
Bihan Wen\textsuperscript{2}\quad
Weisi Lin\textsuperscript{2}\quad
Alex C. Kot\textsuperscript{2,5,6}\quad
Hedvig Kjellstr\"om\textsuperscript{1}\\[4pt]
{\small
\textsuperscript{1}KTH Royal Institute of Technology\quad
\textsuperscript{2}Nanyang Technological University\quad
\textsuperscript{3}Hebei University of Technology}\\
{\small
\textsuperscript{4}Lancaster University\quad
\textsuperscript{5}Shenzhen MSU-BIT University\quad
\textsuperscript{6}VinUniversity}\\[2pt]
{\small \textsuperscript{*}Equal contribution.}
}

\begin{document}
\maketitle

\begin{abstract}
Text-to-motion generation typically relies on dataset-specific skeletal representations, with models trained separately for benchmarks with different joint layouts. 
We investigate whether image diffusion architectures can model 3D human motion through a common image-compatible representation that accommodates heterogeneous skeletal layouts while preserving motion-specific kinematic structure.
To this end, we organize root-centered joint trajectories into topology-informed joint--time images and model global root dynamics through a complementary latent stream. 
A motion autoencoder conditions local-motion decoding on the root latent, while a text-conditioned Scalable Interpolant Transformer (SiT) jointly generates both latent streams, with supervision on decoded 3D joint positions further constraining generation in motion space.
In addition to dataset-specific models, we train a single motion autoencoder on HumanML3D and KIT-ML, followed by a single text-conditioned SiT shared across both datasets, preserving their respective skeletal definitions without skeleton retargeting to a common joint layout.
Experiments on HumanML3D and KIT-ML show that our dataset-specific models achieve state-of-the-art performance, while shared training yields further gains on both benchmarks, demonstrating the benefits of shared modeling across heterogeneous skeletal layouts.
\end{abstract}

\section{Introduction}
\label{sec:introduction}

Text-to-motion generation aims to synthesize realistic 3D human motion from natural-language descriptions~\cite{guo2022generating,petrovich22temos}. 
Recent methods have made substantial progress in motion quality and text–motion alignment~\cite{tevet2023human,zhang2023generating,guo2024momask}, but commonly operate on representations tied to particular skeletal definitions.
Widely used benchmarks such as HumanML3D~\cite{guo2022generating} and KIT-ML~\cite{plappert2016kit} use different skeletal definitions, including different joint counts and indexing conventions. 
In vector-based representations, these differences affect both the vector dimensionality and the correspondence between vector entries and skeletal joints, requiring explicit handling when sharing model inputs and outputs across datasets.
Recent work has explored motion generation across heterogeneous skeletal structures under different conditioning settings, including skeleton-conditioned synthesis~\cite{gat2025anytop}, style- and trajectory-controlled generation~\cite{khani2025unimogen}, and text- and skeleton-conditioned generation~\cite{liu2025texttoanyskeleton}.
Within this broader goal, we investigate whether an image-compatible representation can support shared motion autoencoding and text-conditioned generation while preserving each dataset’s native skeletal definition.

\begin{figure*}[t]
    \centering
    \includegraphics[width=0.9\linewidth]{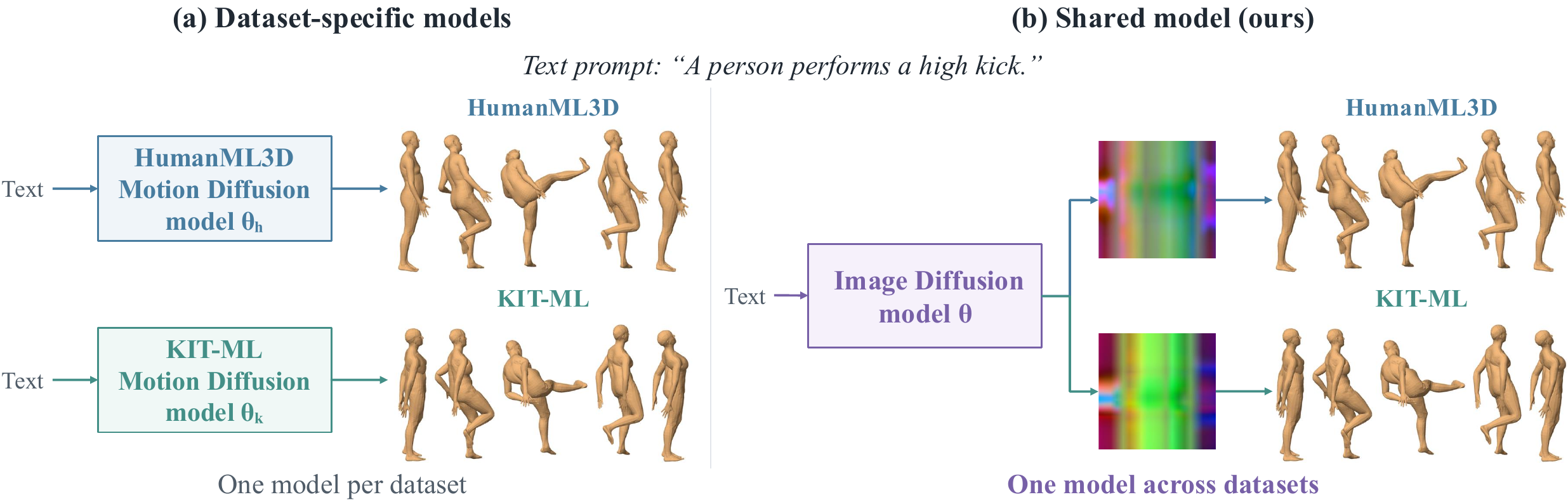}
    \vspace{-0.2cm}
    \caption{\textbf{Dataset-specific versus shared motion modeling.}
    (a) Separate models for HumanML3D and KIT-ML.
    (b) Our method uses a single model across both datasets despite their different skeletal definitions.
    The motion-to-image mapping is detailed in Figure~\ref{fig:motion_to_image}.}
    \vspace{-0.2cm}
    \label{fig:teaser}
\end{figure*}

Root-centered joint trajectories naturally form a joint–time array, with the three spatial coordinates serving as separate channels. 
Mapping this array to a common image resolution makes the spatial dimensions presented to the model independent of the native joint count, providing a basis for shared modeling across datasets (Fig.~\ref{fig:teaser}). 
This image-compatible form also allows motion to be processed by image autoencoding and latent generative architectures~\cite{rombach2022high,peebles2023scalable,ma2024sit}.
However, a common image shape alone does not encode anatomical relationships or capture dependencies between global movement and local articulation. These considerations motivate a representation and modeling framework that explicitly incorporates both.

To address these challenges, we adopt the compact motion representation used in MARDM~\cite{meng2025rethinking}, which demonstrated that root motion and root-centered local joint positions can support strong motion generation. 
We investigate how these two motion components can be organized for image-based generative modeling.
Specifically, we reorder local joints according to body-part organization and kinematic topology and arrange their trajectories into joint–time images, giving the joint axis a meaningful anatomical organization.
Global root dynamics do not share this joint-wise organization and are therefore modeled through a complementary latent stream.
To capture dependencies between global movement and local articulation, our motion autoencoder encodes skeleton images and root motion into separate latent streams and conditions image decoding on the root latent.
A text-conditioned Scalable Interpolant Transformer (SiT) then jointly generates both latent streams from noise, modeling their interaction during generation.
To further constrain generation in motion space, we supervise decoded 3D joint positions.

Beyond dataset-specific training, we use the proposed representation to train a single motion autoencoder across HumanML3D and KIT-ML, followed by a single text-conditioned SiT shared across both datasets. 
A dataset condition specifies the target dataset during latent generation. 
Decoded motions are recovered and supervised according to each dataset’s native skeletal definition, enabling parameter sharing without skeleton retargeting or joint padding.

Experiments on HumanML3D and KIT-ML show that our dataset-specific models achieve state-of-the-art performance, while shared training further improves all six reported metrics on both datasets. 
These results demonstrate that the proposed representation supports both strong dataset-specific generation and shared modeling across heterogeneous skeletal layouts.

Our main contributions are summarized as follows:
\begin{itemize}
    \item We introduce an image-compatible motion representation that organizes root-centered local joint trajectories into topology-informed joint-time images while representing global root dynamics through a complementary stream.
    \item We develop a coupled motion generation framework combining root-conditioned image decoding, joint text-conditioned generation of both latent streams with SiT, and supervision on decoded 3D joint positions.
    \item We enable shared motion autoencoding and text-conditioned generation across HumanML3D and KIT-ML using a single motion autoencoder and a single SiT, while preserving each dataset’s native skeletal definition without skeleton retargeting or joint padding.
\end{itemize}

\section{Related Work}
\textbf{Text-to-Motion Generation.}
Diffusion-based methods generate text-conditioned motion directly in motion space, as in MDM~\cite{tevet2023human} and MotionDiffuse~\cite{zhang2024motiondiffuse}, or in a learned latent space, as in MLD~\cite{chen2023executing}. MARDM~\cite{meng2025rethinking} combines diffusion with masked autoregressive modeling using a compact motion representation.
SALAD~\cite{hong2025salad} introduces skeleton-aware latent diffusion that models relationships among joints, frames, and words for text-driven motion generation and editing.
Another line of work represents motion using discrete tokens.
T2M-GPT~\cite{zhang2023generating} uses autoregressive generation, MoMask~\cite{guo2024momask} combines hierarchical motion quantization with masked prediction, and BAMM~\cite{pinyoanuntapong2024bamm} integrates masked and autoregressive modeling for flexible-length generation and editing. 
Building on discrete motion tokens, MotionLLM~\cite{wu2025motion}  uses a pretrained language model for motion generation, editing, and understanding, while IRG-MotionLLM~\cite{li2026irg} interleaves generation with assessment and refinement.
Complementary to these advances, we investigate an image-compatible motion representation that supports image generative architectures while incorporating skeletal organization and enabling shared modeling across heterogeneous skeletal layouts.

\noindent\textbf{Motion Representations and Image-Compatible Generative Modeling.}
The standard text-to-motion representation~\cite{guo2022generating} combines root motion, joint positions, rotations, velocities, and foot-contact information. 
Following MARDM~\cite{meng2025rethinking}, we use only root motion and root-centered joint positions for image-based generation.
Image-based skeleton representations have also been explored for action recognition. 
Ke et al.~\cite{ke2017new} transform skeleton trajectories into image-based clips encoding spatial relationships for convolutional networks, while SkeleMotion~\cite{caetano2019skelemotion} encodes joint-motion magnitudes and orientations as skeleton images. 
MotionPatches~\cite{yu2024exploring} groups and orders skeletal joints by body part to construct image-like patches, enabling image-pretrained Vision Transformers to encode motion across skeletal structures for text-to-motion retrieval and recognition.
Our work develops this direction for text-conditioned generation: we encode root-centered joint trajectories as topology-informed joint-time images, jointly generate image and root-motion latents, and decode them into 3D motion sequences.

In image synthesis, latent diffusion models~\cite{rombach2022high} generate in an autoencoder’s latent space, while DiT~\cite{peebles2023scalable} and SiT~\cite{ma2024sit} develop Transformer architectures for diffusion and interpolant-based generation. 
We build on these advances to jointly generate image and root-motion latents, using root-conditioned decoding and decoded 3D joint supervision to connect latent generation to motion geometry.

\noindent\textbf{Motion Modeling Across Skeletal Structures.}
Motion retargeting transfers motion from one skeleton to another. Aberman et al.~\cite{aberman2020skeleton} introduce skeleton-aware convolution, pooling, and unpooling operators to embed motions from different skeletal structures into a shared latent space for cross-structural retargeting.
More recent work directly generates motion across heterogeneous skeletal structures. 
AnyTop~\cite{gat2025anytop} incorporates skeletal topology and textual joint descriptions into a diffusion model for skeleton-conditioned synthesis. 
UniMoGen~\cite{khani2025unimogen} dynamically processes character joints within a skeleton-agnostic diffusion model for style- and trajectory-controlled generation.
OmniSkel~\cite{liu2025texttoanyskeleton} generates motion conditioned on text and a target skeleton, using a skeleton-aware motion autoencoder.
Our image-compatible approach maps local joint trajectories to a common image resolution, enabling shared motion autoencoding and generation across datasets with different native joint counts.
We investigate this formulation on HumanML3D and KIT-ML with a single motion autoencoder and a single text-conditioned SiT, using dataset conditioning and dataset-specific recovery to preserve each dataset’s native skeletal definition without retargeting or joint padding.

\section{Method}
\begin{figure*}[t]
    \centering
    \includegraphics[width=0.9\linewidth]{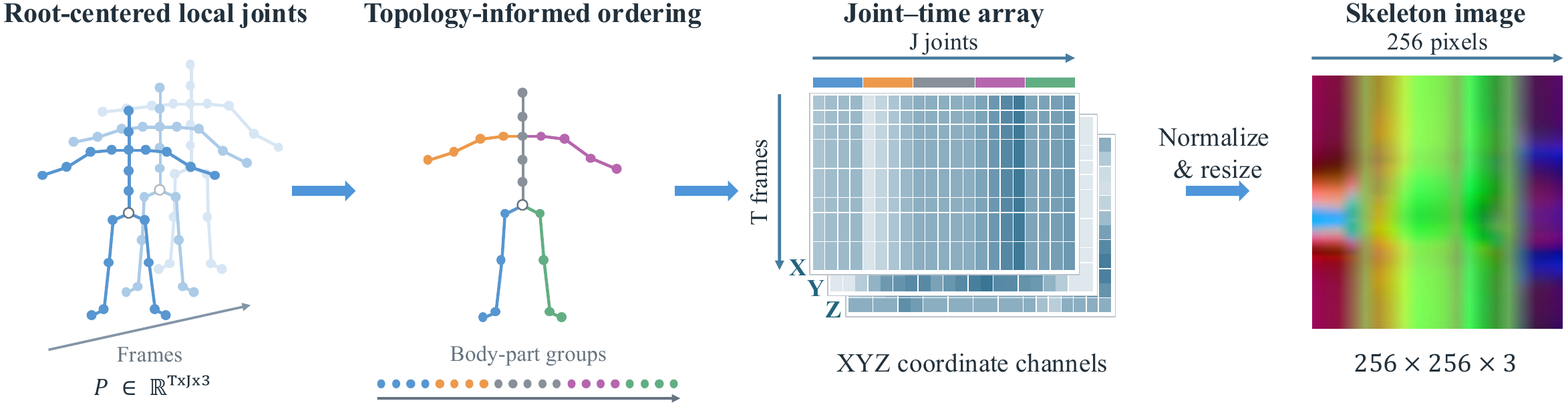}
    \vspace{-0.2cm}
    \caption{\textbf{Topology-informed skeleton image construction.}
    Local joint coordinates are reordered according to body-part grouping and kinematic-chain topology, arranged into a joint–time array with \(x\), \(y\), and \(z\) coordinate channels, and resized to $256 \times 256$.}
    \vspace{-0.2cm}
    \label{fig:motion_to_image}
\end{figure*}

Our framework makes human motion compatible with image generative architectures while modeling dependencies between local articulation and global movement. 
We organize root-centered joint trajectories into topology-informed joint–time images (Fig.~\ref{fig:motion_to_image}) and represent root dynamics through a complementary stream (Sec.~\ref{sec:motion_representation}).
Our motion autoencoder encodes these components into separate latent streams and conditions image decoding on the root latent (Sec.~\ref{sec:motion_autoencoding}).
Building on this latent representation, a text-conditioned SiT jointly generates both streams from noise, modeling their interaction during generation (Sec.~\ref{sec:latent_generation}).
We extend this framework to shared training across HumanML3D and KIT-ML, using a single motion autoencoder and a single SiT without skeleton retargeting (Sec.~\ref{sec:shared_modeling}).
During generator training, supervision on decoded 3D joint positions further constrains generation in motion space (Sec.~\ref{sec:training_objectives}). 
Figure~\ref{fig:framework} illustrates the overall framework.

\subsection{Image-Compatible Motion Representation}
\label{sec:motion_representation}
Following MARDM~\cite{meng2025rethinking}, we represent a $T$-frame motion from dataset $d$ using root features $\mathbf{R}^{(d)} \in \mathbb{R}^{T \times 4}$ and root-centered local joint coordinates $\mathbf{P}^{(d)} \in \mathbb{R}^{T \times J_d \times 3}$.
Here, $J_d$ is the number of non-root joints: 21 for HumanML3D and 20 for KIT-ML. 
We arrange local trajectories into skeleton images and keep root features in a complementary stream.

\noindent\textbf{Topology-informed skeleton images.}
Directly using native joint indices as the image axis would make its spatial organization depend on dataset-specific indexing conventions.
As shown in Fig.~\ref{fig:motion_to_image}, we reorder local joints into a topology-informed layout and arrange their trajectories into a joint--time array.
The ordering follows the left leg, left arm, torso and head, right arm, and right leg.
Within each limb, joints follow the kinematic chain from distal to proximal on the left and from proximal to distal on the right, producing an approximately mirrored layout around the central body.
This arrangement keeps joints within each body part contiguous and gives the joint axis a meaningful anatomical organization.
Section~A of the supplementary material lists the full order for each dataset.
For dataset $d$, let $\mathcal{N}_d$ denote the normalization of local joint coordinates and $\pi_d$ the corresponding joint permutation. We construct
\begin{equation}
\widetilde{\mathbf{P}}^{(d)}_{t,j,:}
=
\left[
\mathcal{N}_d\left(\mathbf{P}^{(d)}\right)
\right]_{t,\pi_d(j),:},
\qquad
\widetilde{\mathbf{P}}^{(d)}
\in \mathbb{R}^{T \times J_d \times 3}.
\end{equation}
Time and joints index the image rows and columns, respectively, while $x$, $y$, and $z$ form the three channels.
We resize this array to a common resolution of $H=W=256$:
\begin{equation}
\mathbf{I}_{\mathrm{mot}}^{(d)}
=
\operatorname{Resize}_{H \times W}
\left(\widetilde{\mathbf{P}}^{(d)}\right)
\in \mathbb{R}^{H \times W \times 3}.
\end{equation}
Across datasets, the same grouping and ordering principles provide a similar coarse anatomical organization along the joint axis.
Together with the common image resolution, this allows the same image encoder and decoder to process motions with different native joint counts, without skeleton retargeting or joint padding.
Decoded images are mapped back to each dataset's native joint-time resolution, joint order, and coordinate scale (Sec.~\ref{sec:motion_autoencoding}).

\noindent\textbf{Complementary root representation.}
We keep the root features outside the skeleton image.
They encode heading changes, two planar root displacements in the heading-aligned coordinate system, and root height above the ground.
These quantities describe global root dynamics and do not share the joint-wise spatial organization of local articulation.
We therefore model $\mathbf{R}^{(d)}$ through a complementary root-motion stream, coupled with local motion during autoencoding and generation.

For brevity, we omit the dataset superscript below when
the formulation applies to both datasets.

\subsection{Root-Conditioned Motion Autoencoding}
\label{sec:motion_autoencoding}
\begin{figure*}[t]
    \centering
    \includegraphics[width=0.95\linewidth]{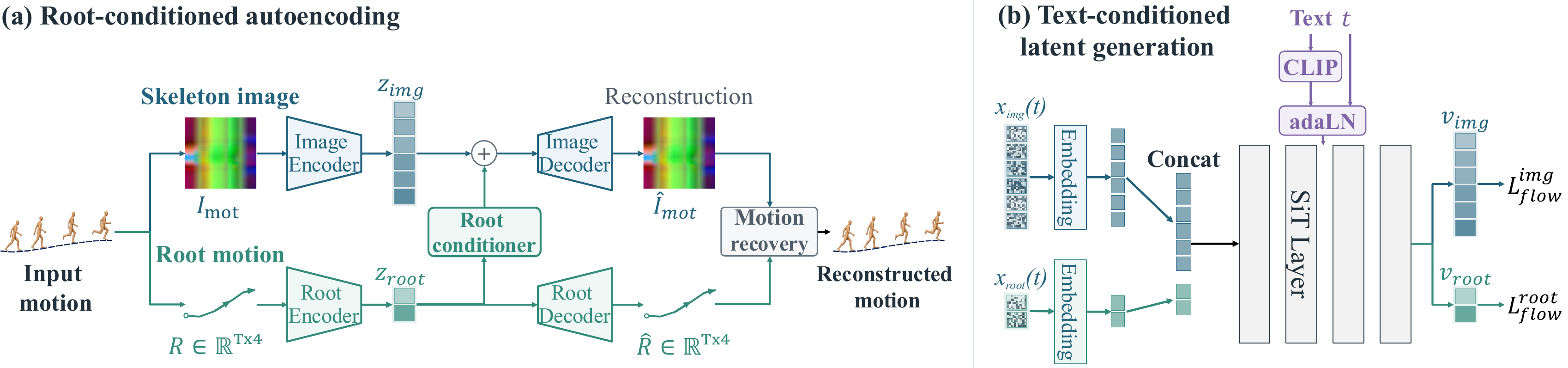}
    \vspace{-0.2cm}
    \caption{\textbf{Overview of the proposed framework.}
    (a) Separate encoders map skeleton images and root motion into two latent spaces.
    Root-conditioned decoding and motion recovery reconstruct the input motion.
    (b) With the autoencoder frozen, noisy image and root latents are tokenized and jointly processed by a text-conditioned SiT to predict their flow-matching velocities.
    }
    \vspace{-0.2cm}
\label{fig:framework}
\end{figure*}

Given the two motion components in Sec.~\ref{sec:motion_representation}, our motion autoencoder uses separate variational encoders to map the skeleton image $\mathbf{I}_{\mathrm{mot}}$ and root sequence $\mathbf{R}$ into an image latent and a root latent:
\begin{equation}
\mathbf z_{\mathrm{img}}
\sim
q_{\phi_{\mathrm{img}}}
\left(
\mathbf z_{\mathrm{img}}
\mid
\mathbf I_{\mathrm{mot}}
\right),
\qquad
\mathbf z_{\mathrm{root}}
\sim
q_{\phi_{\mathrm{root}}}
\left(
\mathbf z_{\mathrm{root}}
\mid
\mathbf R
\right).
\end{equation}
We omit root-feature normalization and denormalization
from the equations for clarity.
The root latent is decoded independently to reconstruct the
global root sequence as $\widehat{\mathbf R} = D_{\mathrm{root}}(\mathbf z_{\mathrm{root}})$.
Although encoded separately, local articulation and global movement remain coupled in natural motion. 
We therefore condition image decoding on the root latent, as shown in Fig.~\ref{fig:framework} (a).
A learned conditioner maps the root latent to the channel dimension and temporal resolution of the image decoder features. 
We broadcast its output along the joint axis and add it to these features before the first upsampling stage.
We denote the complete root-conditioned image decoder by $D_{\mathrm{img}}$:
\begin{equation}
    \hat{\mathbf I}_{\mathrm{mot}}
    =
    D_{\mathrm{img}}\!\left(
        \mathbf z_{\mathrm{img}};
        \mathbf z_{\mathrm{root}}
    \right).
\end{equation}
The root latent thus supports both root-motion reconstruction and local-motion decoding, coupling the two components while retaining separate latent representations.

\noindent\textbf{Motion recovery.}
The decoded skeleton image is resized to the original joint--time resolution, restored to the native joint order, and denormalized to recover the root-centered 3D joint coordinates $\widehat{\mathbf P}$. 
The reconstructed root features are likewise mapped back to their original scale, yielding the recovered motion representation $(\widehat{\mathbf R}, \widehat{\mathbf P})$.

\subsection{Coupled Root--Pose Latent Generation}
\label{sec:latent_generation}

With the motion autoencoder frozen, we use a text-conditioned SiT to jointly generate the image and root latents, as illustrated in Fig.~\ref{fig:framework} (b).

\noindent\textbf{Latent interpolation.}
For each stream $s\in\{\mathrm{img},\mathrm{root}\}$, we use the encoder's posterior mean as the clean latent $\mathbf z_s$ and independently sample Gaussian noise $\boldsymbol{\epsilon}_s\sim\mathcal N(\mathbf 0,\mathbf I)$.
Both streams share the same generative time $t\in[0,1]$, which is distinct from the motion-frame index.
We use a linear interpolant with its corresponding velocity target:
\begin{equation}
\mathbf x_s(t)
= t\mathbf z_s+(1-t)\boldsymbol{\epsilon}_s,
\qquad
\mathbf u_s
= \mathbf z_s-\boldsymbol{\epsilon}_s.
\end{equation}

\noindent\textbf{Text-conditioned joint latent modeling.}
Given a text description, a frozen text encoder provides both a global text condition $\mathbf c_{\mathrm{text}}$ and token-level text features.
The global condition is combined with the timestep embedding to modulate the SiT blocks through adaptive layer normalization, while the token-level features are projected into text prefix tokens $\mathbf h_{\mathrm{text}}$ for joint self-attention with the motion tokens.

To construct the joint input sequence, we tokenize the current image and root states separately and project the resulting motion tokens to the Transformer's hidden dimension.
The text prefix tokens use the same hidden dimension,
allowing all three token sequences to be concatenated:
\begin{equation}
\mathbf H(t)
=
\left[
\mathbf h_{\mathrm{text}};
\operatorname{Tok}_{\mathrm{root}}
\left(
\mathbf x_{\mathrm{root}}(t)
\right);
\operatorname{Tok}_{\mathrm{img}}
\left(
\mathbf x_{\mathrm{img}}(t)
\right)
\right].
\end{equation}
Each token sequence includes its own learnable positional embeddings before concatenation; implementation details are provided in Sec.~B.2 of the supplementary material.
Shared Transformer blocks process this sequence through joint self-attention, allowing the two motion streams to exchange information with each other and the text tokens.
Separate output heads map the corresponding Transformer features back to their respective latent spaces and predict a velocity field for each stream:
\begin{equation}
\left(
\mathbf v_{\theta}^{\mathrm{img}},
\mathbf v_{\theta}^{\mathrm{root}}
\right)
=
F_{\theta}
\left(
\mathbf H(t),
t;
\mathbf c_{\mathrm{text}}
\right).
\end{equation}
Each prediction therefore depends on both current latent states and the text description, coupling local articulation and global root dynamics throughout generation.

\noindent\textbf{Simultaneous sampling.}
At inference, we initialize both streams with independent Gaussian noise and evolve them jointly from $t=0$ to $t=1$ using the predicted velocities.
The generated root latent reconstructs the root features and conditions local-motion reconstruction from the generated image latent.
This couples global movement and local articulation during both latent generation and motion decoding.

\subsection{Shared Modeling Across Skeletal Layouts}
\label{sec:shared_modeling}

The common image representation enables shared modeling across HumanML3D and KIT-ML despite their different native joint counts and skeletal definitions.
We first train a single motion autoencoder on both datasets, sharing the image and root encoders, decoders, and root conditioner.
Each motion is normalized using dataset-specific training statistics and mapped to a fixed-resolution skeleton image with a complementary root sequence.
Decoded outputs are restored to the corresponding dataset's native joint--time resolution, joint order, and coordinate scale.
No skeleton retargeting or joint padding is required.

We then freeze the shared autoencoder and train a single SiT on paired latents from both datasets.
To specify the target dataset during generation, we augment the global condition with a learnable dataset embedding $\mathbf e_d$:
\begin{equation}
\mathbf c_d
=
\tau(t)
+
\psi(\mathbf c_{\mathrm{text}})
+
\mathbf e_d,
\end{equation}
where $\tau(t)$ is the timestep embedding and $\psi$ projects the global text condition $\mathbf{c}_{\mathrm{text}}$ to the SiT hidden dimension.
The combined condition modulates the shared Transformer blocks through adaptive layer normalization, while text prefix conditioning remains unchanged.

For classifier-free guidance, we jointly drop the text condition and replace the dataset ID with a null ID during training.
At inference, the conditional branch uses the target dataset embedding, while the unconditional branch uses a learned null embedding $\mathbf e_0$.
Generated latents are decoded by the shared autoencoder and recovered in the target dataset's native skeletal representation.

\begin{table*}[t]
    \centering
    \caption{
        Quantitative comparison on HumanML3D using the MARDM evaluation protocol.
        \textbf{Best} and \underline{second-best} results are marked.
        Shared models are jointly trained on HumanML3D and KIT-ML.
        Confidence intervals are included where reported.
    }
    \vspace{-0.2cm}
    \label{tab:humanml3d_comparison}
    \setlength{\tabcolsep}{4pt}
    \renewcommand{\arraystretch}{1.12}

    \resizebox{0.9\textwidth}{!}{%
    \begin{tabular}{lcccccc}
        \toprule

        \multirow{2}{*}{Method}
        & \multicolumn{3}{c}{R-Precision $\uparrow$}
        & \multirow{2}{*}{FID $\downarrow$}
        & \multirow{2}{*}{Matching $\downarrow$}
        & \multirow{2}{*}{CLIP-score $\uparrow$} \\

        \cmidrule(lr){2-4}
        & Top-1 & Top-2 & Top-3 & & & \\
        \midrule

        \multicolumn{7}{l}{\textit{Motion Discrete / Token-based}} \\

        T2M-GPT~\cite{zhang2023generating}
        & $0.470^{\pm .003}$
        & $0.659^{\pm .002}$
        & $0.758^{\pm .002}$
        & $0.335^{\pm .003}$
        & $3.505^{\pm .017}$
        & $0.607^{\pm .005}$ \\

        MMM~\cite{pinyoanuntapong2024mmm}
        & $0.487^{\pm .003}$
        & $0.683^{\pm .002}$
        & $0.782^{\pm .001}$
        & $0.132^{\pm .004}$
        & $3.359^{\pm .009}$
        & $0.635^{\pm .003}$ \\

        MoMask~\cite{guo2024momask}
        & $0.490^{\pm .004}$
        & $0.687^{\pm .003}$
        & $0.786^{\pm .003}$
        & $0.116^{\pm .006}$
        & $3.353^{\pm .010}$
        & $0.637^{\pm .003}$ \\

        BAMM~\cite{pinyoanuntapong2024bamm}
        & $0.480^{\pm .003}$
        & $0.671^{\pm .003}$
        & $0.772^{\pm .003}$
        & $0.147^{\pm .005}$
        & --
        & $0.630^{\pm .001}$ \\

        MoMask++~\cite{NEURIPS2025_9028b8a3}
        & $0.492^{\pm .003}$
        & $0.683^{\pm .003}$
        & $0.782^{\pm .002}$
        & $0.108^{\pm .007}$
        & --
        & $0.634^{\pm .001}$ \\

        \midrule

        \multicolumn{7}{l}{\textit{Motion Diffusion / Flow-based}} \\

        MDM-50Step~\cite{tevet2023human}
        & $0.440^{\pm .007}$
        & $0.636^{\pm .006}$
        & $0.742^{\pm .004}$
        & $0.518^{\pm .032}$
        & $3.640^{\pm .028}$
        & $0.578^{\pm .003}$ \\

        MotionDiffuse~\cite{zhang2024motiondiffuse}
        & $0.450^{\pm .006}$
        & $0.641^{\pm .005}$
        & $0.753^{\pm .005}$
        & $0.778^{\pm .005}$
        & $3.490^{\pm .023}$
        & $0.606^{\pm .004}$ \\

        MLD~\cite{chen2023executing}
        & $0.461^{\pm .004}$
        & $0.651^{\pm .004}$
        & $0.750^{\pm .003}$
        & $0.431^{\pm .014}$
        & $3.445^{\pm .019}$
        & $0.610^{\pm .003}$ \\

        ReMoDiffuse~\cite{zhang2023remodiffuse}
        & $0.468^{\pm .003}$
        & $0.653^{\pm .003}$
        & $0.754^{\pm .005}$
        & $0.883^{\pm .021}$
        & $3.414^{\pm .020}$
        & $0.621^{\pm .003}$ \\

        MARDM-DDPM-XL~\cite{meng2025rethinking}
        & $0.492^{\pm .006}$
        & $0.690^{\pm .005}$
        & $0.790^{\pm .005}$
        & $0.116^{\pm .004}$
        & $3.349^{\pm .010}$
        & $0.637^{\pm .005}$ \\

        MARDM-SiT-XL~\cite{meng2025rethinking}
        & $0.500^{\pm .004}$
        & $0.695^{\pm .003}$
        & $0.795^{\pm .003}$
        & $0.114^{\pm .007}$
        & $3.270^{\pm .009}$
        & $0.642^{\pm .002}$ \\

        MLD++~\cite{motionlcm_v2}
        & $0.500^{\pm .003}$
        & $0.691^{\pm .002}$
        & $0.789^{\pm .001}$
        & $2.027^{\pm .021}$
        & $3.220^{\pm .008}$
        & $0.639^{\pm .002}$ \\

        MotionLCM-V2~\cite{motionlcm_v2}
        & $0.501^{\pm .002}$
        & $0.693^{\pm .002}$
        & $0.790^{\pm .002}$
        & $2.267^{\pm .023}$
        & {$3.192^{\pm .009}$}
        & $0.640^{\pm .003}$ \\

        \midrule

        \multicolumn{7}{l}{\textit{Motion LLM-based}} \\

        MotionLLM~\cite{wu2025motion}
        & $0.470$
        & $0.664$
        & $0.771$
        & $0.240$
        & $3.377$
        & -- \\

        IRG-MotionLLM (S1 + IRMoGen)~\cite{li2026irg}
        & $0.507$
        & $0.696$
        & $0.791$
        & $0.161$
        & $3.234$
        & -- \\

        IRG-MotionLLM (S3 + IRMoGen)~\cite{li2026irg}
        & $0.509$
        & $0.701$
        & $0.802$
        & $0.259$
        & $\underline{3.166}$
        & -- \\

        \midrule

        \multicolumn{7}{l}{\textit{Image Diffusion / Flow-based}} \\
        
        \textbf{Ours (SiT-L)}
        & $\underline{0.540^{\pm .004}}$
        & $\underline{0.727^{\pm .002}}$
        & $\underline{0.818^{\pm .002}}$
        & $\underline{0.089^{\pm .005}}$
        & $3.226^{\pm .010}$
        & $\underline{0.654^{\pm .001}}$ \\

        \textbf{Ours (SiT-L, Shared)}
        & $\mathbf{0.546^{\pm .002}}$
        & $\mathbf{0.733^{\pm .002}}$
        & $\mathbf{0.822^{\pm .002}}$
        & $\mathbf{0.082^{\pm .004}}$
        & $\mathbf{3.154^{\pm .006}}$
        & $\mathbf{0.659^{\pm .001}}$ \\

        \bottomrule
    \end{tabular}%
    }
    \vspace{-0.2cm}
\end{table*}

\subsection{Training Objectives}
\label{sec:training_objectives}
We first train the motion autoencoder with reconstruction and KL losses, then freeze it and train the SiT with latent flow matching and decoded joint-position supervision.

\noindent\textbf{Autoencoder training.}
Each latent stream is trained with a reconstruction loss and KL regularization.
For the image stream, reconstruction combines mean absolute and mean squared errors between the input and reconstructed skeleton images; for the root stream, it uses an L1 loss on root features.
We additionally recover root-centered local 3D joint positions from the reconstructed skeleton images and apply an L1 reconstruction loss in the native joint order and coordinate scale.
The autoencoder objective is
\begin{equation}
\mathcal L_{\mathrm{VAE}}
=
\mathcal L_{\mathrm{img\text{-}rec}}
+
\mathcal L_{\mathrm{root\text{-}rec}}
+
\mathcal L_{\mathrm{local\text{-}rec}}
+
\beta_{\mathrm{img}}\mathcal L_{\mathrm{KL}}^{\mathrm{img}}
+
\beta_{\mathrm{root}}\mathcal L_{\mathrm{KL}}^{\mathrm{root}}.
\end{equation}
Here, $\mathcal L_{\mathrm{KL}}^{s}$ regularizes each variational posterior toward a standard Gaussian prior,
with weights $\beta_{\mathrm{img}}$ and $\beta_{\mathrm{root}}$ for the image and root streams, respectively.
Through root-conditioned decoding, the image and local-joint reconstruction losses also propagate gradients to the root encoder.
For shared training, we compute the objective for each dataset and aggregate it as $\mathcal L_{\mathrm{VAE}}^{\mathrm{shared}} =\sum_d w_d\mathcal L_{\mathrm{VAE}}^{(d)}$, where $w_d$ denotes the dataset weight and $\sum_d w_d=1$.

\noindent\textbf{Latent generator training.}
With the autoencoder frozen, we train the SiT to predict the velocity targets.
For each stream $s\in\{\mathrm{img},\mathrm{root}\}$, we minimize $\mathcal L_{\mathrm{flow}}^{s} = \mathbb E[\operatorname{MSE}(\mathbf v_{\theta}^{s},\mathbf u_s)]$.
We average over paired motion--text samples, interpolation times, and Gaussian noise.

To constrain generation in motion space, we supervise local 3D joint positions decoded from predicted clean latents. Under the linear interpolant, these estimates are
\begin{equation}
\widehat{\mathbf z}_s
=
\mathbf x_s(t)
+
(1-t)\mathbf v_{\theta}^{s}.
\end{equation}
We decode the estimated image latent conditioned on the estimated root latent and recover local joint coordinates:
\begin{equation}
    \hat{\mathbf P}_{\theta}
    =
    \mathcal G\!\left(
        D_{\mathrm{img}}\!\left(
            \hat{\mathbf z}_{\mathrm{img}};
            \hat{\mathbf z}_{\mathrm{root}}
        \right)
    \right),
\end{equation}
where $D_{\mathrm{img}}$ is the root-conditioned image decoder, and $\mathcal G$ is the image-to-motion recovery operation. 
For shared modeling, we use the corresponding dataset-specific operator $\mathcal G_d$.
The position loss is $\mathcal L_{\mathrm{pos}}
= \mathbb E[ \operatorname{MAE}(\widehat{\mathbf P}_{\theta},\mathbf P)]$.
Although the autoencoder is frozen, gradients flow through the decoder and root conditioner to both predicted latents.

Combining latent flow matching with decoded geometry supervision, the final SiT objective is
\begin{equation}
\mathcal L_{\mathrm{SiT}}
=
\mathcal L_{\mathrm{flow}}^{\mathrm{img}}
+
\lambda_{\mathrm{root}}
\mathcal L_{\mathrm{flow}}^{\mathrm{root}}
+
\lambda_{\mathrm{pos}}
\mathcal L_{\mathrm{pos}}.
\end{equation}
Here, $\lambda_{\mathrm{root}}$ and $\lambda_{\mathrm{pos}}$ weight the root-stream flow-matching loss and decoded joint-position loss, respectively.
For shared training, samples from both datasets use their corresponding dataset conditions.
We combine the per-dataset objectives as $\mathcal{L}_{\mathrm{SiT}}^{\mathrm{shared}} = \sum_d w_d \mathcal{L}_{\mathrm{SiT}}^{(d)}$.

\section{Experiments}
\subsection{Experimental Setup}

\textbf{Datasets.}
We evaluate on HumanML3D~\cite{guo2022generating} and KIT-ML~\cite{plappert2016kit}, following the standard T2M preprocessing protocol~\cite{guo2022generating}.
HumanML3D contains 14,616 motion sequences paired with 44,970 textual descriptions, while KIT-ML contains 3,911 motion sequences with 6,278 textual annotations.
Motions are processed at 20 FPS for HumanML3D and 12.5 FPS for KIT-ML.
We follow the standard train/val/test splits (approximately 80\%, 5\%, and 15\%) and use mirroring augmentation with corresponding text adjustments.
Following MARDM~\cite{meng2025rethinking}, we use four-dimensional root features and root-centered local joint positions, yielding 67-dimensional and 64-dimensional per-frame representations for HumanML3D and KIT-ML.

\noindent\textbf{Evaluation Metrics.}
We follow the MARDM evaluation protocol~\cite{meng2025rethinking}, using its pretrained evaluators for the 67-dimensional HumanML3D and 64-dimensional KIT-ML representations.
For both reconstruction and generation, we use the ground-truth valid window length $T \leq 196$ as the target output length.
Both local-joint and root-feature outputs are resampled to $T$ before evaluator normalization, then zero-padded to 196 frames and passed to the evaluator together with the valid length $T$.
The T2M-style evaluator~\cite{guo2022generating} measures text-motion alignment through R-Precision (Top-1, Top-2, and Top-3; higher is better) and Matching Score (lower is better).
We also report Fr\'echet Inception Distance (FID; lower is better) between generated and real motion embeddings to assess distributional similarity.
Finally, we use MARDM's CLIP-based motion-text evaluator to report CLIP-score (higher is better) as a measure of text--motion alignment.
We repeat evaluation 20 times and report mean results with 95\% confidence intervals.

\begin{table*}[t]
    \centering
    \caption{
        Quantitative comparison on KIT-ML using the
        MARDM evaluation protocol.
    }
    \vspace{-0.2cm}
    \label{tab:kitml_comparison}
    \setlength{\tabcolsep}{4pt}
    \renewcommand{\arraystretch}{1.12}

    \resizebox{0.8\textwidth}{!}{%
    \begin{tabular}{lcccccc}
        \toprule

        \multirow{2}{*}{Method}
        & \multicolumn{3}{c}{R-Precision $\uparrow$}
        & \multirow{2}{*}{FID $\downarrow$}
        & \multirow{2}{*}{Matching $\downarrow$}
        & \multirow{2}{*}{CLIP-score $\uparrow$} \\

        \cmidrule(lr){2-4}
        & Top-1 & Top-2 & Top-3 & & & \\
        \midrule

        \multicolumn{7}{l}{\textit{Motion Discrete / Token-based}} \\

        T2M-GPT~\cite{zhang2023generating}
        & $0.359^{\pm .007}$
        & $0.553^{\pm .007}$
        & $0.690^{\pm .013}$
        & $0.593^{\pm .053}$
        & $3.765^{\pm .046}$
        & $0.651^{\pm .005}$ \\

        MMM~\cite{pinyoanuntapong2024mmm}
        & $0.363^{\pm .005}$
        & $0.569^{\pm .006}$
        & $0.724^{\pm .006}$
        & $0.478^{\pm .034}$
        & $3.629^{\pm .028}$
        & $0.660^{\pm .003}$ \\

        MoMask~\cite{guo2024momask}
        & $0.369^{\pm .005}$
        & $0.588^{\pm .005}$
        & $0.731^{\pm .005}$
        & $0.411^{\pm .026}$
        & $3.577^{\pm .021}$
        & $0.669^{\pm .002}$ \\

        \midrule

        \multicolumn{7}{l}{\textit{Motion Diffusion / Flow-based}} \\

        MDM~\cite{tevet2023human}
        & $0.333^{\pm .012}$
        & $0.561^{\pm .009}$
        & $0.689^{\pm .009}$
        & $0.585^{\pm .043}$
        & $4.002^{\pm .033}$
        & $0.605^{\pm .007}$ \\

        MotionDiffuse~\cite{zhang2024motiondiffuse}
        & $0.344^{\pm .009}$
        & $0.536^{\pm .007}$
        & $0.658^{\pm .007}$
        & $3.845^{\pm .087}$
        & $4.167^{\pm .054}$
        & $0.626^{\pm .006}$ \\

        MLD~\cite{chen2023executing}
        & $0.351^{\pm .007}$
        & $0.536^{\pm .007}$
        & $0.658^{\pm .007}$
        & $0.492^{\pm .047}$
        & $3.746^{\pm .044}$
        & $0.646^{\pm .006}$ \\

        ReMoDiffuse~\cite{zhang2023remodiffuse}
        & $0.356^{\pm .004}$
        & $0.572^{\pm .007}$
        & $0.706^{\pm .009}$
        & $1.725^{\pm .053}$
        & $3.735^{\pm .036}$
        & $0.665^{\pm .005}$ \\

        MARDM-DDPM-XL~\cite{meng2025rethinking}
        & $0.375^{\pm .006}$
        & $0.597^{\pm .008}$
        & $0.739^{\pm .006}$
        & $0.340^{\pm .020}$
        & $3.489^{\pm .018}$
        & $0.681^{\pm .003}$ \\

        MARDM-SiT-XL~\cite{meng2025rethinking}
        & $0.387^{\pm .006}$
        & $0.610^{\pm .006}$
        & $0.749^{\pm .006}$
        & $0.242^{\pm .014}$
        & $3.374^{\pm .019}$
        & $0.692^{\pm .002}$ \\

        \midrule

        \multicolumn{7}{l}{\textit{Image Diffusion / Flow-based}} \\
        
        \textbf{Ours (SiT-L)}
        & \underline{$0.395^{\pm .004}$}
        & \underline{$0.620^{\pm .007}$}
        & \underline{$0.758^{\pm .004}$}
        & \underline{$0.171^{\pm .010}$}
        & \underline{$3.331^{\pm .020}$}
        & \underline{$0.696^{\pm .002}$} \\
        
        \textbf{Ours (SiT-L, Shared)}
        & $\mathbf{0.405^{\pm .006}}$
        & $\mathbf{0.634^{\pm .006}}$
        & $\mathbf{0.770^{\pm .007}}$
        & $\mathbf{0.145^{\pm .010}}$
        & $\mathbf{3.292^{\pm .011}}$
        & $\mathbf{0.708^{\pm .002}}$ \\

        \bottomrule
    \end{tabular}%
    }
    \vspace{-0.3cm}
\end{table*}

\noindent\textbf{Implementation Details.}
We train both dataset-specific models and a shared model on HumanML3D and KIT-ML using the two-stage procedure in Sec.~\ref{sec:training_objectives}.
The motion autoencoder uses an ImageNet-pretrained ResNet-18 image encoder~\cite{he2016deep,deng2009imagenet} and a 1D convolutional root encoder.
With the autoencoder frozen, we train SiT-L~\cite{ma2024sit} to jointly generate both latent streams, using a frozen CLIP ViT-B/32 text encoder~\cite{radford2021learning} for conditioning.
For shared training, we mix HumanML3D and KIT-ML samples at a 3:1 ratio, approximately reflecting their relative dataset sizes, and use corresponding dataset weights of 0.75 and 0.25.
At inference, we jointly sample both latent streams with a 50-step Euler solver and classifier-free guidance~\cite{ho2022classifier}, using guidance scales of 5.5 for HumanML3D and 4.5 for KIT-ML.
All experiments are implemented in PyTorch and run on NVIDIA A100 GPUs.
Further architecture and training details are provided in Sec.~B of the supplementary material.

\begin{figure*}[t]
    \centering
    \includegraphics[width=0.9\textwidth]{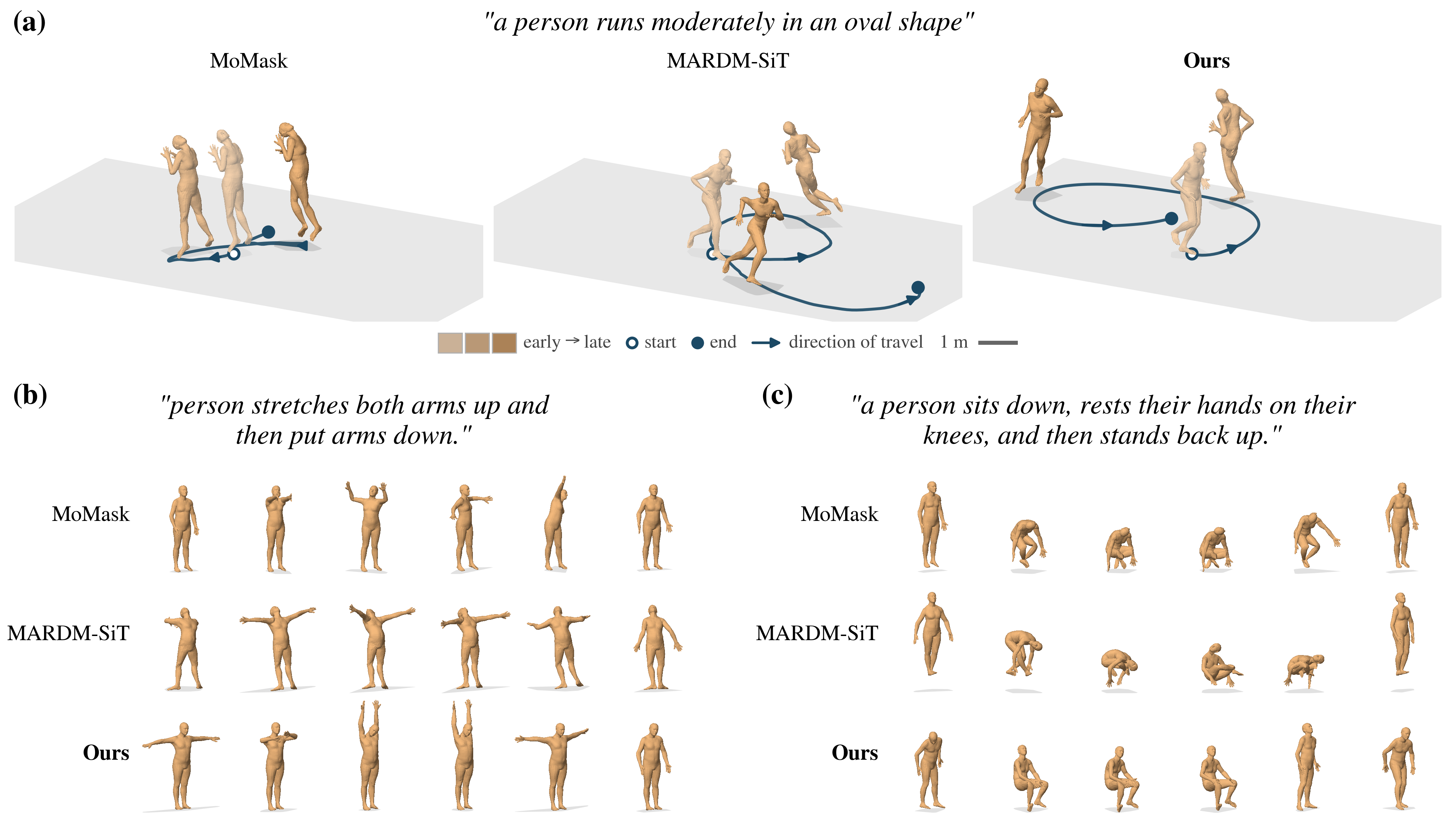}
    \vspace{-0.2cm}
    \caption{\textbf{Qualitative comparisons on HumanML3D.}
    (a) Motion snapshots with root trajectories, where lighter-to-darker meshes indicate temporal progression.
    (b)--(c) Motion sequences shown chronologically from left to right.
    }
    \vspace{-0.4cm}
    \label{fig:qualitative}
\end{figure*}

\subsection{Comparison with Existing Methods}
\label{sec:comparison}

We compare with representative text-to-motion approaches on HumanML3D and KIT-ML, including discrete/token-based, diffusion/flow-based, and LLM-based methods.

\noindent\textbf{Dataset-specific generation.}
On HumanML3D, our dataset-specific SiT-L model achieves the highest R-Precision at all three retrieval levels, the lowest FID, and the highest CLIP-score among the dataset-specific methods reported in Table~\ref{tab:humanml3d_comparison}.
It also obtains a competitive Matching Score of 3.226.
Using the same root-motion and root-centered joint-position features as MARDM-SiT-XL~\cite{meng2025rethinking}, our model achieves higher Top-1 R-Precision (0.540 vs.\ 0.500) and lower FID (0.089 vs.\ 0.114).
On KIT-ML, our model leads the dataset-specific methods on all six metrics, as shown in Table~\ref{tab:kitml_comparison}.
In particular, our model achieves lower FID (0.171 vs.\ 0.242) and higher Top-1 R-Precision (0.395 vs.\ 0.387) than MARDM-SiT-XL.

\noindent\textbf{Shared modeling across datasets.}
We train a single motion autoencoder on HumanML3D and KIT-ML, then freeze it to train a shared SiT-L generator.
Compared with our dataset-specific models, shared training improves all six reported metrics on both benchmarks.
On HumanML3D, FID decreases from 0.089 to 0.082 and Top-1 R-Precision increases from 0.540 to 0.546.
On KIT-ML, FID decreases from 0.171 to 0.145 and Top-1 R-Precision increases from 0.395 to 0.405.
The shared model achieves the best reported results across all six metrics in both tables.
These results support the effectiveness of shared modeling across the two skeletal layouts using our common image representation, without skeleton retargeting or joint padding.

\subsection{Qualitative Results}
Fig.~\ref{fig:qualitative} compares our method with MoMask and MARDM-SiT on three text prompts.
Our method captures running along an oval path in (a), overhead arm extension followed by lowering in (b), and sitting with hands resting on the knees followed by standing in (c).
These examples illustrate the model's ability to capture both global movement and local articulation specified by text.
The supplementary material includes a video of generated motions.

\subsection{Ablation Studies}
\label{sec:ablation}

\begin{table}[t]
    \centering
    \caption{Model ablations on HumanML3D.}
    \vspace{-0.3cm}
    \label{tab:ablation_design}

    \setlength{\tabcolsep}{1.5pt}
    \renewcommand{\arraystretch}{0.90}

    \resizebox{0.95\linewidth}{!}{%
    \begin{tabular}{@{}lcccc@{}}
        \toprule
        \multirow{2}{*}{Setting}
        & \multirow{2}{*}{FID $\downarrow$}
        & \multicolumn{3}{c}{R-Precision $\uparrow$} \\
        \cmidrule(lr){3-5}
        & & Top-1 & Top-2 & Top-3 \\
        \midrule

        \multicolumn{5}{l}{\textit{Model scaling}} \\
        SiT-S
        & $0.139^{\pm .004}$
        & $0.527^{\pm .004}$
        & $0.713^{\pm .002}$
        & $0.804^{\pm .002}$ \\
        SiT-B
        & $0.106^{\pm .006}$
        & $0.533^{\pm .003}$
        & $0.717^{\pm .002}$
        & $0.810^{\pm .002}$ \\
        SiT-L
        & $\mathbf{0.089^{\pm .005}}$
        & $\mathbf{0.540^{\pm .004}}$
        & $\mathbf{0.727^{\pm .002}}$
        & $\mathbf{0.818^{\pm .002}}$ \\

        \midrule
        \multicolumn{5}{l}{\textit{Joint ordering}} \\
        Original order
        & $0.116^{\pm .007}$
        & $0.526^{\pm .002}$
        & $0.713^{\pm .002}$
        & $0.806^{\pm .002}$ \\
        Fixed random order
        & $0.121^{\pm .011}$
        & $0.524^{\pm .006}$
        & $0.711^{\pm .002}$
        & $0.805^{\pm .002}$ \\
        Topology-informed
        & $\mathbf{0.106^{\pm .006}}$
        & $\mathbf{0.533^{\pm .003}}$
        & $\mathbf{0.717^{\pm .002}}$
        & $\mathbf{0.810^{\pm .002}}$ \\

        \midrule
        \multicolumn{5}{l}{\textit{ImageNet initialization}} \\
        w/o ImageNet
        & $0.145^{\pm .006}$
        & $0.517^{\pm .003}$
        & $0.705^{\pm .003}$
        & $0.800^{\pm .002}$ \\
        w/ ImageNet
        & $\mathbf{0.106^{\pm .006}}$
        & $\mathbf{0.533^{\pm .003}}$
        & $\mathbf{0.717^{\pm .002}}$
        & $\mathbf{0.810^{\pm .002}}$ \\

        \midrule
        \multicolumn{5}{l}{\textit{Root-conditioned decoding}} \\
        w/o condition
        & $0.116^{\pm .005}$
        & $0.525^{\pm .004}$
        & $0.714^{\pm .003}$
        & $0.807^{\pm .003}$ \\
        w/ condition
        & $\mathbf{0.106^{\pm .006}}$
        & $\mathbf{0.533^{\pm .003}}$
        & $\mathbf{0.717^{\pm .002}}$
        & $\mathbf{0.810^{\pm .002}}$ \\

        \midrule
        \multicolumn{5}{l}{\textit{Dataset conditioning (Shared)}} \\
        w/o condition
        & $0.278^{\pm .015}$
        & $0.515^{\pm .005}$
        & $0.694^{\pm .008}$
        & $0.781^{\pm .004}$ \\
        w/ condition
        & $\mathbf{0.093^{\pm .008}}$
        & $\mathbf{0.536^{\pm .004}}$
        & $\mathbf{0.726^{\pm .007}}$
        & $\mathbf{0.818^{\pm .004}}$ \\

        \bottomrule
    \end{tabular}%
    }
    \vspace{-0.3cm}
\end{table}

We conduct ablations on HumanML3D using SiT-B with $\lambda_{\mathrm{pos}} = \lambda_{\mathrm{root}} = 0.25$ unless otherwise specified.
The dataset-conditioning ablation uses shared models trained on HumanML3D and KIT-ML; all other ablations use dataset-specific training. 
Additional loss-weight ablations are reported in Sec.~C.2 of the supplementary material.

\noindent\textbf{Model scaling.}
We compare SiT-S, SiT-B, and SiT-L to assess generator capacity.
As shown in Table~\ref{tab:ablation_design}, FID decreases from 0.139 to 0.106 and 0.089, while all three R-Precision metrics improve.
These results show that our framework benefits from increased generator capacity.

\noindent\textbf{Joint ordering.}
We compare three joint orderings for the skeleton image: the original HumanML3D order, a fixed random permutation, and our topology-informed order.
As shown in Table~\ref{tab:ablation_design}, the topology-informed order achieves the lowest FID (0.106, versus 0.116 for the original order and 0.121 for the random permutation) and the highest R-Precision at all three retrieval levels.
Without the autoencoder, resizing the local joint trajectories to image resolution and back yields a local MPJPE of 3.5 and 3.7~mm for 128- and 196-frame windows, respectively, compared with 12.0 and 12.1~mm for the original HumanML3D order.
Section~C.1 of the supplementary material reports the full resize and VAE reconstruction comparisons.

\noindent\textbf{ImageNet initialization.}
We compare image encoders with and without ImageNet initialization.
Table~\ref{tab:ablation_design} shows that using pretrained weights reduces FID from 0.145 to 0.106 and improves Top-1 R-Precision from 0.517 to 0.533.
These results show that ImageNet initialization benefits motion generation with our image-compatible representation.

\noindent\textbf{Root-conditioned decoding.}
We evaluate the effect of conditioning image decoding on the root latent.
As reported in Table~\ref{tab:ablation_design}, removing this conditioning increases FID from 0.106 to 0.116 and decreases Top-1 R-Precision from 0.533 to 0.525.
This supports the benefit of incorporating root-motion context into local-motion decoding.

\noindent\textbf{Dataset conditioning.}
We compare shared SiT-B models with and without the dataset condition.
The results in Table~\ref{tab:ablation_design} show that adding this condition reduces FID from 0.278 to 0.093 and improves Top-3 R-Precision from 0.781 to 0.818.
Without it, shared training performs worse than the dataset-specific baseline with FID 0.106.
These results suggest that explicit dataset conditioning helps distinguish two motion distributions within the shared representation.

\section{Conclusion}
\label{sec:conclusion}
In this work, we show that image generative architectures can effectively model text-conditioned 3D human motion through a structured, image-compatible representation.
Our framework combines topology-informed joint--time images with a complementary root-motion stream, coupling the two through root-conditioned autoencoding and joint SiT generation.
Supervision on decoded 3D joint positions further constrains generation in motion space.
This representation also enables a single motion autoencoder and a single SiT to model HumanML3D and KIT-ML jointly, preserving their native skeletal definitions without retargeting.
Experiments demonstrate state-of-the-art dataset-specific results and further improvements through shared training on both benchmarks.
These findings highlight the potential of image-compatible representations for motion generation across heterogeneous skeletal layouts.

\subsection*{LLM Usage Statement}
We used ChatGPT (OpenAI) to assist in two aspects of this work: writing and debugging experiment code (e.g., scripts for training, evaluation, and analysis), and improving the language and clarity of the manuscript.
The research ideas, method design, and conclusions are entirely our own.
All AI-assisted outputs, including text revisions and code suggestions, were carefully reviewed by the authors.
The authors take full responsibility for the content of this paper.

\section*{Acknowledgements}
We gratefully acknowledge access to the Berzelius computing resource at the National Supercomputer
Centre, provided by the Knut and Alice Wallenberg Foundation. This work was supported by the
Wallenberg AI, Autonomous Systems and Software Program (WASP).

{
    \small
    \bibliographystyle{ieeenat_fullname}
    \bibliography{main}
}

\end{document}


\clearpage
\setcounter{page}{1}
\maketitlesupplementary

\renewcommand{\thesection}{\Alph{section}}                    
\renewcommand{\thesubsection}{\thesection.\arabic{subsection}} 
\renewcommand{\thetable}{S\arabic{table}}                     
\renewcommand{\thefigure}{S\arabic{figure}}                   

In this supplementary material, we first describe the topology-informed joint orderings used for HumanML3D and KIT-ML (Section~\ref{app:joint_ordering}). 
We then provide implementation details for the motion autoencoder and SiT generator
(Section~\ref{app:implementation}).
We quantify the error from resizing local joint trajectories to image resolution and back, evaluate VAE reconstruction quality, and examine the effect of the loss weights (Section~\ref{app:supp_additional_experiments}).
Finally, we provide additional qualitative comparisons in an accompanying video, with visualization details described in Section~\ref{sec:supp_qualitative}.

Throughout this supplement, section numbers (e.g., Sec.~3.1) refer to the main paper, while lettered references (e.g., Sec.~C.1) refer to this document.

\section{Topology-informed Joint Ordering}
\label{app:joint_ordering}

We organize local joints into five consecutive body-part groups: left leg, left arm, torso and head, right arm, and right leg.
Within the left limbs, joints are ordered from distal to proximal; within the right limbs, they are ordered from proximal to distal. 
The central group follows the torso toward the head.
This arrangement keeps each body part contiguous along the joint axis and provides a similar coarse anatomical organization across datasets.

Table~\ref{tab:joint_ordering} lists the joint names and native indices in the order used to construct the skeleton images.
The root joint has native index 0 in both datasets and is excluded from the local joint array, leaving 21 local joints for HumanML3D and 20 for KIT-ML.
We retain each dataset's native root definition: the pelvis in HumanML3D and the root joint in KIT-ML.

\begin{table}[h]
    \centering
    \caption{
        Body-part grouping and joint ordering for skeleton image
        construction. Rows follow the group order along the joint
        axis; joints within each cell are listed in traversal order.
        Parentheses indicate zero-based native joint indices,
        before root removal.
    }
    \label{tab:joint_ordering}

    \small
    \setlength{\tabcolsep}{5pt}
    \renewcommand{\arraystretch}{1.2}

    \resizebox{0.9\linewidth}{!}{%
    \begin{tabularx}{\linewidth}{
        @{}
        >{\raggedright\arraybackslash}p{0.14\linewidth}
        >{\raggedright\arraybackslash}X
        >{\raggedright\arraybackslash}X
        @{}
    }
        \toprule
        Body part & HumanML3D & KIT-ML \\
        \midrule

        Left leg
        & Foot (10), ankle (7), knee (4), hip (1)
        & Toe (20), mid-foot (19), ankle (18),
          knee (17), hip (16) \\

        \addlinespace
        Left arm
        & Wrist (20), elbow (18), shoulder (16), collar (13)
        & Wrist (10), elbow (9), shoulder (8) \\

        \addlinespace
        Torso and head
        & Spine1 (3), spine2 (6), spine3 (9),
          neck (12), head (15)
        & Pelvis spine (1), thorax (2), lower neck (3), upper neck (4) \\

        \addlinespace
        Right arm
        & Collar (14), shoulder (17), elbow (19), wrist (21)
        & Shoulder (5), elbow (6), wrist (7) \\

        \addlinespace
        Right leg
        & Hip (2), knee (5), ankle (8), foot (11)
        & Hip (11), knee (12), ankle (13),
          mid-foot (14), toe (15) \\

        \bottomrule
    \end{tabularx}
    }
\end{table}

The datasets differ in both joint counts and anatomical partitioning.
Each leg contains four local joints in HumanML3D and five in KIT-ML, whereas each arm contains four and three, respectively;
the torso-and-head groups contain five and four.

Concatenating the ordered groups in Table~\ref{tab:joint_ordering} specifies the joint-axis ordering.
The table uses native skeleton indices, whereas the permutation $\pi_d$ in Sec.~3.1 acts on the local joint array after root removal.
During recovery, the inverse permutation restores the dataset-specific local joint order.
The grouping does not establish a one-to-one joint correspondence between datasets and requires neither skeleton retargeting nor joint padding.

\section{Implementation Details}
\label{app:implementation}

\subsection{Motion Autoencoder Architecture}

\textbf{Image encoder.}
Skeleton images have a resolution of $256 \times 256$.
The image encoder uses the stem and the first three residual stages of an ImageNet-pretrained ResNet-18 backbone.
The resulting image latent has 16 channels and a spatial resolution of $16 \times 16$.

\noindent\textbf{Image decoder.}
The image decoder first maps the 16-channel image latent to 256-channel features using a $3 \times 3$ convolution with padding 1, followed by group normalization with 32 groups, a SiLU activation, and a residual block.
Root conditioning is added to these features before the first upsampling stage.
Four successive stages then increase the spatial resolution from $16 \times 16$ to $256 \times 256$, with channel dimensions $256 \rightarrow 128 \rightarrow 64 \rightarrow 32 \rightarrow 16$.
Each stage consists of $2\times$ nearest-neighbor upsampling, a $3 \times 3$ convolution, group normalization, a SiLU activation, and a residual block.
Group normalization uses 32 groups in the first three stages and 16 groups in the final stage.
The output head applies a $3 \times 3$ convolution with 16 output channels, group normalization with 16 groups, a SiLU activation, and a final $3 \times 3$ convolution to reconstruct the three-channel skeleton image.
Together with the root conditioner, this complete decoding process is denoted by $D_{\mathrm{img}}(\mathbf z_{\mathrm{img}}; \mathbf z_{\mathrm{root}})$ in Sec.~3.2.

\noindent\textbf{Root encoder.}
The four-channel root sequence is first linearly interpolated to 16 temporal positions, then processed by two length-preserving 1D convolutions with kernel size 3 and channel dimensions $4 \rightarrow 128 \rightarrow 128$.
Each convolution is followed by group normalization with 32 groups and a SiLU activation.
Separate kernel-size-one convolutional heads predict the posterior mean and log variance, yielding a root latent of size $8 \times 16$ (channels $\times$ temporal positions).
Temporal compression is performed by the initial interpolation, while the convolutional layers preserve the sequence length.

\noindent\textbf{Root decoder.}
The root decoder applies two length-preserving 1D convolutions with kernel size 3 and channel dimensions $8 \rightarrow 128 \rightarrow 128$.
Each convolution is followed by group normalization with 32 groups and a SiLU activation.
A final kernel-size-one convolution maps the features to four root channels, which are then linearly interpolated to the original sequence length $T$.

\noindent\textbf{Root conditioner.}
The root conditioner uses two kernel-size-one 1D convolutions with channel dimensions $8 \rightarrow 256 \rightarrow 256$.
Group normalization with 32 groups and a SiLU activation are applied between the two convolutions.
The resulting features are temporally aligned with the image decoding features by linear interpolation when needed, then broadcast along the joint axis.
In our configuration, this produces a conditioning tensor with 256 channels and a spatial resolution of $16 \times 16$.
It is added after the image decoder's initial feature-mapping block, before any upsampling.

\subsection{SiT Architecture and Text Conditioning}

\textbf{Model configurations.}
We use SiT-L for the main experiments and examine SiT-S and SiT-B in the model-size ablation.
Table~\ref{tab:sit_configs} summarizes their configurations.
All variants use an attention head dimension of 64, an MLP ratio of 4, and an image-latent patch size of $1 \times 1$.
Parameter counts refer to the trainable parameters of our instantiated generators.

\begin{table}[h]
    \centering
    \caption{SiT configurations and trainable parameter counts.}
    \label{tab:sit_configs}
    \resizebox{0.95\linewidth}{!}{%
    \begin{tabular}{lrrrr}
        \toprule
        Model & Depth & Hidden dimension & Heads & Parameters \\
        \midrule
        SiT-S & 12 & 384  & 6  & 33.4M  \\
        SiT-B & 12 & 768  & 12 & 132.3M \\
        SiT-L & 24 & 1024 & 16 & 461.3M \\
        \bottomrule
    \end{tabular}
    }
\end{table}

\noindent\textbf{Latent tokenization and prediction.}
The image and root latents are tokenized into 256 image tokens and 16 root tokens, respectively.
Each image token corresponds to one spatial position in the image latent, while each root token corresponds to one temporal position in the root latent. Separate learned projections map both token sequences to the Transformer hidden dimension $d$.

Independent learnable positional embeddings are added to the image, root, and text prefix tokens before concatenation.
The corresponding embedding tables have shapes $256 \times d$, $16 \times d$, and $8 \times d$, and are initialized from a zero-mean Gaussian distribution with a standard deviation of 0.02.
The sequences are concatenated in the order of text, root, and image, yielding a joint sequence of 280 tokens.
No additional positional embedding is applied after concatenation.
Shared Transformer blocks process this sequence through joint self-attention.
Separate output heads map the image and root token features back to their respective latent spaces to predict the two velocity fields.

\noindent\textbf{Text and dataset conditioning.}
Text descriptions are encoded using a frozen CLIP ViT-B/32 text encoder. 
Its pooled embedding provides the global text condition, which is combined with the interpolation-time embedding to condition the Transformer blocks through adaptive layer normalization.
To also retain token-level text information, we uniformly select eight valid text tokens and project their embeddings to the Transformer hidden dimension.
After adding their learnable positional embeddings, these tokens are prepended to the motion tokens as described above.
For shared modeling, a learned dataset embedding is additionally included in the global condition (Sec.~3.4).

\subsection{Training and Sampling}

We follow the two-stage procedure in Sec.~3.5, first training the motion autoencoder and then freezing it to train SiT.
We report the settings for HumanML3D, KIT-ML, and shared training separately in Tables~\ref{tab:ae_training_settings} and~\ref{tab:sit_training_settings}.
All batch sizes are effective batch sizes across GPUs.

\noindent\textbf{Autoencoder training.}
We train the motion autoencoder on sliding windows of 128 frames for HumanML3D and 64 frames for KIT-ML.
Shared training retains these dataset-specific window lengths.
We use AdamW~\citep{loshchilov2018decoupled} with a learning-rate warmup followed by cosine decay.
The KL regularization weights are also warmed up at the beginning of training, with the warmup durations and target weights listed in Table~\ref{tab:ae_training_settings}.

\begin{table}[t]
    \centering
    \caption{Motion autoencoder training settings.}
    \label{tab:ae_training_settings}
    \renewcommand{\arraystretch}{1.15}
    \resizebox{\linewidth}{!}{%
    \begin{tabular}{lccc}
        \toprule
        Setting & HumanML3D & KIT-ML & Shared \\
        \midrule
        \multirow{2}{*}{Training window (frames)}
            & \multirow{2}{*}{128}
            & \multirow{2}{*}{64}
            & 128 (HumanML3D) \\
            & & & 64 (KIT-ML) \\
        \addlinespace[3pt]
        Effective batch size
            & 256 & 512 & 256 \\
        Epochs
            & 100 & 50 & 100 \\
        Optimizer
            & \multicolumn{3}{c}{AdamW} \\
        Learning rate
            & \multicolumn{3}{c}{$10^{-3}$} \\
        Weight decay
            & \multicolumn{3}{c}{0.05} \\
        Learning-rate schedule
            & \multicolumn{3}{c}{Cosine} \\
        LR warmup (epochs)
            & 1 & 5 & 1 \\
        KL warmup (epochs)
            & 2 & 5 & 2 \\
        Image KL weight $\beta_{\mathrm{img}}$
            & \multicolumn{3}{c}{$7\times10^{-4}$} \\
        Root KL weight $\beta_{\mathrm{root}}$
            & \multicolumn{3}{c}{$1\times10^{-4}$} \\
        Training precision
            & \multicolumn{3}{c}{FP32} \\
        \bottomrule
    \end{tabular}
    }
\end{table}

\noindent\textbf{SiT training.}
With the motion autoencoder frozen, we use its posterior means as clean latent targets and train SiT on motions of up to 196 frames.
We optimize SiT using AdamW with a constant learning rate, without learning-rate warmup or weight decay. Training uses automatic mixed precision (FP16) without gradient clipping.
We maintain an exponential moving average (EMA) of the generator parameters and use the EMA weights for evaluation.
Conditioning dropout enables classifier-free guidance, while the root-stream and decoded joint-position loss weights control their contributions to the training objective.
The corresponding settings are listed in Table~\ref{tab:sit_training_settings}.

\begin{table}[t]
    \centering
    \caption{SiT training settings.}
    \label{tab:sit_training_settings}
    \renewcommand{\arraystretch}{1.15}
    \resizebox{\linewidth}{!}{%
    \begin{tabular}{lccc}
        \toprule
        Setting & HumanML3D & KIT-ML & Shared \\
        \midrule
        Maximum motion length (frames)
            & \multicolumn{3}{c}{196} \\
        Effective batch size
            & 256 & 16 & 256 \\
        Epochs
            & \multicolumn{3}{c}{2000} \\
        Optimizer
            & \multicolumn{3}{c}{AdamW} \\
        Adam betas $(\beta_1,\beta_2)$
            & \multicolumn{3}{c}{$(0.9,0.999)$} \\
        Learning rate
            & \multicolumn{3}{c}{$10^{-4}$} \\
        Learning-rate schedule
            & \multicolumn{3}{c}{Constant} \\
        EMA decay
            & \multicolumn{3}{c}{0.9999} \\
        Training precision
            & \multicolumn{3}{c}{AMP (FP16)} \\
        Conditioning dropout probability
            & \multicolumn{3}{c}{0.1} \\
        Root loss weight $\lambda_{\mathrm{root}}$
            & \multicolumn{3}{c}{0.25} \\
        Position loss weight $\lambda_{\mathrm{pos}}$
            & \multicolumn{3}{c}{0.25} \\
        \bottomrule
    \end{tabular}
    }
\end{table}

\noindent\textbf{Shared training.}
In both stages, each mixed batch contains HumanML3D and KIT-ML samples at a 3:1 ratio, approximately reflecting their relative dataset sizes.
We compute the objectives separately for each dataset and aggregate them using weights $w_{\mathrm{HumanML3D}}=0.75$ and $w_{\mathrm{KIT\text{-}ML}}=0.25$, as defined in Sec.~3.5.
During SiT training, each sample uses its corresponding
dataset condition. We jointly drop the text and dataset conditions with probability 0.1, following Sec.~3.4.

\noindent\textbf{Sampling.}
At inference, we initialize both latent streams with independent Gaussian noise and jointly integrate them using a 50-step Euler solver.
We apply classifier-free guidance~\citep{ho2022classifier} with scales of 5.5 for HumanML3D and 4.5 for KIT-ML. 
For the shared model, the conditional branch uses the target dataset embedding, while the unconditional branch uses the learned null embedding. 
The generated latents are decoded and recovered to the target dataset's native motion representation, following Sec.~3.2.

\section{Additional Experiments}
\label{app:supp_additional_experiments}

The generation ablation in Sec.~4.4 shows that the topology-informed joint order achieves the lowest FID among the three orderings on HumanML3D. 
Here, we examine whether its advantage is already evident when local joint trajectories are resized to an image and back, and whether it persists after VAE reconstruction (Sec.~\ref{sec:supp_reconstruction}). 
We then examine how the weights on decoded joint supervision and root flow matching affect text-to-motion generation (Sec.~\ref{app:loss_weight_ablations}).

\subsection{Motion Representation and Reconstruction Fidelity}
\label{sec:supp_reconstruction}

\noindent\textbf{Effect of image resizing.}
We isolate the error introduced by image resizing by omitting the autoencoder. 
For each joint ordering, we resize the root-centered joint trajectories to a $256 \times 256$ image using bilinear interpolation, resize them back to the original temporal and joint dimensions, and restore the native joint order before measuring the error.
Because interpolation mixes neighboring columns along the joint axis, the choice of which joints are placed next to one another matters.
As shown in Table~\ref{tab:supp_resize}, our topology-informed order yields a local MPJPE of 3.5~mm for 128-frame windows and 3.7~mm for 196-frame windows. 
At both lengths, the error is below one-third of that obtained with the original HumanML3D order.

\begin{table}[t]
    \centering
    \caption{Effect of joint ordering on the image resize round trip
    on HumanML3D. No autoencoder is used.}
    \label{tab:supp_resize}
    \setlength{\tabcolsep}{6pt}
    \begin{tabular}{@{}lcc@{}}
        \toprule
        & \multicolumn{2}{c}{Local MPJPE (mm) $\downarrow$} \\
        \cmidrule(lr){2-3}
        Joint ordering & 128 frames & 196 frames \\
        \midrule
        Original HumanML3D order & 12.0 & 12.1 \\
        Fixed random permutation &  9.6 &  9.8 \\
        Topology-informed order  & \textbf{3.5} & \textbf{3.7} \\
        \bottomrule
    \end{tabular}
\end{table}

\begin{table}[t]
    \centering
    \caption{Motion reconstruction on the HumanML3D test set.
    Baseline results are from Table~A1 of MARDM~\citep{meng2025rethinking}.
    We follow the same protocol and report pelvis-aligned MPJPE.
    Values are means with 95\% confidence intervals.
    Bold and underlined values indicate the best and second-best
    results, respectively.}
    \label{tab:supp_vae_reconstruction}

    \setlength{\tabcolsep}{5pt}
    \resizebox{\linewidth}{!}{%
    \begin{tabular}{@{}lcc@{}}
        \toprule
        Method & FID $\downarrow$ & MPJPE (mm) $\downarrow$ \\
        \midrule
        \multicolumn{3}{@{}l}{\textit{Reported in MARDM}} \\
        VQ-VAE (T2M-GPT)~\citep{zhang2023generating}
        & $0.0810 \pm 0.0010$ & $72.600 \pm 0.001$ \\
        RVQ-VAE (MoMask)~\citep{guo2024momask}
        & $0.0290 \pm 0.0010$ & $31.500 \pm 0.001$ \\
        VAE (MLD)~\citep{chen2023executing}
        & $0.0230 \pm 0.0010$ & $13.700 \pm 0.001$ \\
        MARDM AE~\citep{meng2025rethinking}
        & $\mathbf{0.0040} \pm 0.0010$ & $\mathbf{1.000} \pm 0.001$ \\
        \midrule
        \multicolumn{3}{@{}l}{\textit{Our VAE by joint ordering}} \\
        Original HumanML3D order
        & $0.0474 \pm 0.0003$ & $8.554 \pm 0.006$ \\
        Fixed random permutation
        & $0.0390 \pm 0.0003$ & $7.357 \pm 0.003$ \\
        Topology-informed order
        & $\underline{0.0181} \pm 0.0002$
        & $\underline{5.788} \pm 0.002$ \\
        \bottomrule
    \end{tabular}%
    }
\end{table}

\noindent\textbf{VAE reconstruction.}
We evaluate our VAE with three joint orderings on the HumanML3D test set following the MARDM protocol~\citep{meng2025rethinking}.
As shown in Table~\ref{tab:supp_vae_reconstruction}, the topology-informed order yields the lowest reconstruction FID (0.0181) and pelvis-aligned MPJPE (5.788~mm) among our three variants.
These values are also lower than those reported for the VQ-VAE, RVQ-VAE, and VAE baselines.
Although our VAE ranks second to MARDM AE on both reconstruction metrics, our dataset-specific SiT-L outperforms MARDM-SiT-XL on both generation FID and Top-1 R-Precision (Table~1 of the main paper).
On HumanML3D, our SiT-L reduces generation FID from 0.114 to \textbf{0.089} and increases Top-1 R-Precision from 0.500 to \textbf{0.540}.

\subsection{Loss-Weight Ablations}
\label{app:loss_weight_ablations}

We study the effect of the decoded geometry weight $\lambda_{\mathrm{pos}}$ and the root flow-matching weight $\lambda_{\mathrm{root}}$ on text-to-motion generation.
All experiments in this subsection use dataset-specific SiT-B models trained on HumanML3D. We vary one weight at a time while keeping the other at its default value of 0.25.

\begin{table}[t]
    \centering
    \caption{Decoded geometry weight ablation on HumanML3D,
    with $\lambda_{\mathrm{root}}=0.25$.}
    \label{tab:ablation_pos_weight}

    \setlength{\tabcolsep}{2pt}
    \renewcommand{\arraystretch}{1.05}
    \resizebox{\linewidth}{!}{%
    \begin{tabular}{@{}lcccc@{}}
        \toprule
        \multirow{2}{*}{$\lambda_{\mathrm{pos}}$}
        & \multirow{2}{*}{FID $\downarrow$}
        & \multicolumn{3}{c}{R-Precision $\uparrow$} \\
        \cmidrule(lr){3-5}
        & & Top-1 & Top-2 & Top-3 \\
        \midrule
        0 (w/o loss)
        & $0.133^{\pm .006}$
        & $0.520^{\pm .003}$
        & $0.709^{\pm .003}$
        & $0.802^{\pm .002}$ \\
        0.25 (default)
        & $\mathbf{0.106^{\pm .006}}$
        & $\mathbf{0.533^{\pm .003}}$
        & $\mathbf{0.717^{\pm .002}}$
        & $\mathbf{0.810^{\pm .002}}$ \\
        0.50
        & $0.112^{\pm .005}$
        & $0.529^{\pm .003}$
        & $0.715^{\pm .002}$
        & $0.809^{\pm .003}$ \\
        0.75
        & $0.114^{\pm .004}$
        & $0.529^{\pm .002}$
        & $0.716^{\pm .002}$
        & $0.808^{\pm .003}$ \\
        1.00
        & $0.118^{\pm .005}$
        & $0.528^{\pm .002}$
        & $0.713^{\pm .003}$
        & $0.805^{\pm .002}$ \\
        \bottomrule
    \end{tabular}%
    }
\end{table}

\begin{table}[t]
    \centering
    \caption{Root flow-matching weight ablation on HumanML3D,
    with $\lambda_{\mathrm{pos}}=0.25$.}
    \label{tab:ablation_root_weight}

    \setlength{\tabcolsep}{2pt}
    \renewcommand{\arraystretch}{1.05}
    \resizebox{\linewidth}{!}{%
    \begin{tabular}{@{}lcccc@{}}
        \toprule
        \multirow{2}{*}{$\lambda_{\mathrm{root}}$}
        & \multirow{2}{*}{FID $\downarrow$}
        & \multicolumn{3}{c}{R-Precision $\uparrow$} \\
        \cmidrule(lr){3-5}
        & & Top-1 & Top-2 & Top-3 \\
        \midrule
        0.25 (default)
        & $\mathbf{0.106^{\pm .006}}$
        & $\mathbf{0.533^{\pm .003}}$
        & $\mathbf{0.717^{\pm .002}}$
        & $\mathbf{0.810^{\pm .002}}$ \\
        0.50
        & $0.108^{\pm .005}$
        & $0.532^{\pm .003}$
        & $\mathbf{0.717^{\pm .003}}$
        & $\mathbf{0.810^{\pm .002}}$ \\
        0.75
        & $0.109^{\pm .004}$
        & $0.531^{\pm .003}$
        & $0.716^{\pm .002}$
        & $0.808^{\pm .002}$ \\
        1.00
        & $0.111^{\pm .006}$
        & $0.532^{\pm .003}$
        & $0.716^{\pm .002}$
        & $0.809^{\pm .002}$ \\
        \bottomrule
    \end{tabular}%
    }
\end{table}

\noindent\textbf{Decoded geometry weight.}
We examine the effect of supervision on decoded 3D joint positions by varying $\lambda_{\mathrm{pos}}$.
As shown in Table~\ref{tab:ablation_pos_weight}, removing this loss increases FID from $0.106$ to $0.133$ and decreases Top-1 R-Precision from $0.533$ to $0.520$.
Among the tested values, $\lambda_{\mathrm{pos}}=0.25$ achieves the best mean FID and all three mean R-Precision scores.
Larger weights still outperform the setting without this loss, but do not improve upon the default setting.
These results support the benefit of decoded geometry supervision for downstream motion generation.

\noindent\textbf{Root flow-matching weight.}
We vary $\lambda_{\mathrm{root}}$ to examine the balance between the image and root flow-matching objectives.
Table~\ref{tab:ablation_root_weight} shows that performance varies only slightly over the tested range.
The default value $\lambda_{\mathrm{root}}=0.25$ achieves the lowest mean FID and highest mean Top-1 R-Precision, while matching the best reported Top-2 and Top-3 R-Precision.
These results support our default choice of
$\lambda_{\mathrm{root}}=0.25$.

\section{Additional Qualitative Comparisons}
\label{sec:supp_qualitative}

The accompanying video (\texttt{16440\_video\_S1.mp4}) presents nine selected examples from the HumanML3D test set, comparing MoMask~\citep{guo2024momask}, MARDM-SiT~\citep{meng2025rethinking}, and our method from left to right.
For each example, all methods receive the same text prompt and target motion length, and one generated sample is shown per method.
To show both global movement and local articulation, the upper row uses a fixed camera, while the lower row tracks the root horizontally.
For selected locomotion examples, we additionally display root trajectories and starting positions.

For visualization, we fit SMPL meshes to each method's generated joints using the same fitting procedure and a fixed body shape shared across methods, and play the resulting sequences at 20 fps in real time.
The fitting uses no temporal smoothing loss, and neither the generated motions nor the fitted meshes undergo temporal filtering or foot locking.
Only the human motions are rendered; objects mentioned or implied by the prompts, such as chairs and balance beams, are omitted.

{
    \small
    \bibliographystyle{ieeenat_fullname}
    \bibliography{main}
}